\pdfoutput=1

\documentclass[11pt]{article}

\usepackage[]{ACL2023}

\usepackage{times}
\usepackage{latexsym}

\usepackage[T1]{fontenc}

\usepackage[utf8]{inputenc}

\usepackage{microtype}

\usepackage{inconsolata}

\usepackage{amsmath,verbatim}
\usepackage{balance}
\usepackage{makecell}
\usepackage{algorithm}
\usepackage{algpseudocode}
\usepackage{amssymb}
\usepackage{xcolor}
\usepackage{graphicx}
\usepackage{booktabs}
\usepackage{pgfplots,pgfplotstable}
\pgfplotsset{compat=1.18}
\usepackage{tikz, multirow}
\usepackage{orcidlink}

\title{Byte Pair Encoding Is All You Need For Automatic Bengali Speech Recognition}

\author{Ahnaf Mozib Samin \orcidlink{0000-0002-5736-4628}\\
  Faculty of Arts, University of Groningen, The Netherlands \\
  Faculty of Information and Communication Technology, University of Malta, Malta \\
  \texttt{ahnaf.samin.22@um.edu.mt} \\}

\begin{document}
\maketitle
\begin{abstract}
  Byte pair encoding (BPE) emerges as an effective tokenization method for tackling the out-of-vocabulary (OOV) challenge in various natural language and speech processing tasks. Recent research highlights the dependency of BPE subword tokenization's efficacy on the morphological nature of the language, particularly in languages rich in inflectional morphology, where fewer BPE merges suffice for generating highly productive tokens. Motivated by this, our study empirically identifies the optimal number of BPE tokens for Bengali, a language known for its morphological complexity, thus enhancing out-of-distribution automatic speech recognition (ASR) performance. Experimental evaluation reveals that an excessively high number of BPE tokens can lead to overfitting, while approximately 500-1000 tokens result in superior OOV performance. Furthermore, we conduct a comparative analysis of BPE with character-based and unigram-based tokenization methods. By introducing BPE tokenization to Bengali ASR, we achieve a substantial reduction in the word error rate (WER) from 66.44\% in our character-based baseline system to 63.80\% on the LB-ASRTD eval set and from 46.34\% to 42.80\% on the SHRUTI eval set, both of which include out-of-distribution data.
\end{abstract}

\section{Introduction}
\label{sec:intro}
The performance of an automatic speech recognition system is contingent on its core components including acoustic feature extraction, mapping acoustic features to tokens using the Gaussian Mixture Model/Hidden Markov Model (GMM-HMM) or neural networks, and language model-based rescoring of the outputs from connectionist temporal classification (CTC), etc \cite{graves2006connectionist}. As for segmentation, either word-level or subword-level tokens can be modeled to build ASR systems. However, word-level modeling faces a challenge due to the vast number of words in a language, which exceeds the typical vocabulary size of an ASR system. As a result, word-level modeling is susceptible to the out-of-vocabulary (OOV) problem \cite{livescu2012subword}. An OOV word refers to a word that was not encountered during model training but appears during the inference phase.

Subword units are known as effective solutions to tackle the OOV issue in natural language processing (NLP) \cite{sennrich-etal-2016-neural}. Examples of subword units include phonemes, characters, unigrams, and byte pair encoding (BPE) tokens, etc \cite{kudo2018subword, gage1994new, sennrich-etal-2016-neural}. Phonemes represent the smallest units of sound, and after training with phonemes, a model can infer new words. Creating a lexicon that maps each word to its corresponding phonemes, however, requires domain expertise as well as a substantial amount of time and effort for manual annotation. Character-based ASR models are easier to develop since mapping between words and their corresponding characters can be done using an automated way. Furthermore, training a model with a limited number of characters in a language is more computationally efficient than training a word-based model. Unigram language modeling is another segmentation technique that removes tokens based on language model perplexity, initially applied in machine translation \cite{kudo2018subword}.

BPE subword tokenization is first utilized in neural machine translation (NMT) and has gained widespread usage due to their ability to handle OOV words effectively \cite{gage1994new, sennrich-etal-2016-neural}. Subsequent studies implement BPE-based subword modeling in the speech processing domain \cite{synnaeve2020end, yusuyin2023investigation}. For BPE, the number of merge operations determines the number of generated tokens/subwords. From the work of \citet{gutierrez2023languages} on 47 diverse languages, it has been found that in languages characterized by extensive inflectional morphology, there is a tendency to generate highly productive subwords during the initial merging steps. Conversely, in languages with limited inflectional morphology, idiosyncratic subwords tend to play a more prominent role. Therefore, the characteristics of subwords and the required number of merges in BPE tokenization are contingent upon the morphological nature of the respective language. Moreover, an empirical study is conducted by incrementally increasing the BPE merges going from characters to words \cite{gutierrez2021characters}. The authors found that around 200 BPE merges result in the most similar distribution across different languages.

Since the optimal number of BPE merges cannot be universally determined for different languages with varied types of morphology \cite{gutierrez2023languages}, in this study, we empirically determine the number of BPE merges needed for a highly inflectional language—Bengali to achieve superior out-of-distribution ASR performance. Bengali is an Indo-Aryan language spoken in Bangladesh and India and poses challenges to the development of robust ASR systems due to its intricate morphological forms and inadequate research \cite{ali2008morphological, samin2021deep}. Subsequently, we compare the results of BPE-based tokenization with segmentation approaches based on characters and unigrams by performing a cross-dataset evaluation, aiming to understand their effectiveness for handling out-of-distribution data. This investigation exploiting different subword modeling approaches is conducted for the first time for Bengali ASR, to the best of our knowledge.

The rest of the paper is structured as follows: Section \ref{sec:background} provides a comprehensive background study on BPE and unigram language modeling, along with a review of related work in the speech processing domain. In Section \ref{sec:method}, we describe the methodology of our experiments. Section \ref{sec:experiment_setup} provides details about the experiment setup. Results are discussed in Section \ref{sec:results}. Section \ref{sec:conclusion} concludes the article and outlines the future directions.

\section{Background}
\label{sec:background}
\subsection{Byte pair encoding}
Byte pair encoding is a data compression algorithm, which was applied in NMT in 2016 \cite{gage1994new, sennrich-etal-2016-neural}. 

\begin{algorithm}
\caption{Byte-pair encoding \cite{gage1994new, sennrich-etal-2016-neural, bostram-durrett-2020-byte}}\label{alg:cap}
\begin{algorithmic}
\State $S \gets $ set of strings (Approx. 40k Bengali words)
\State $n \gets $ target vocab size
\Procedure{BPE}{$S,n$}
\State $V \gets $ all unique characters in $S$
\While{|$V| < n$}
\State \textbf{Step 1} Merge tokens $a$ and $b$, where 
\State \hspace{1.1cm} $a, b \in V$ and represent the most
\State \hspace{1.1cm} frequent bigram in $S$
\State \textbf{Step 2} Create a new token $ab$ by 
\State \hspace{1.1cm} concatenating $a$ and $b$
\State \textbf{Step 3} Add $ab$ to $V$
\State \textbf{Step 4} Replace each bigram occurrence 
\State \hspace{1.1cm} of $a, b$ tokens in $S$ with $ab$
\EndWhile
\State \textbf{return} $V$
\EndProcedure
\end{algorithmic}
\end{algorithm}

BPE algorithm takes a set of strings $S$ and aims to create a vocabulary $V$ with a target size of $n$. It iteratively merges the most frequent bigram in $S$ into a new token, updating the vocabulary and replacing occurrences of the merged tokens in the original strings. The algorithm continues until the vocabulary size reaches the desired target size $n$ and returns the final vocabulary $V$.

\begin{algorithm}
\caption{Unigram LM \cite{kudo2018subword, bostram-durrett-2020-byte}}\label{alg:cap}
\begin{algorithmic}
\State $S \gets $ set of strings (Approx. 40k Bengali words)
\State $n \gets $ target vocab size
\Procedure{Unigram}{$S,n$}
\State $V \gets $ all substrings occurring more than
\State \hspace{0.7cm} once in $S$ (not crossing words)
\While{|$V| > n$}
\State Build the unigram language model $\theta$ 
\State with $S$
\For{$t$ in $V$}
\State $L_t$ $\leftarrow$ $pplx_{\theta}(S)$ $-$ $pplx_{\theta^{'}}(S)$
\State where $\theta^{'}$ is the LM without token t
\EndFor
\State Remove min(|$V$| $-$ $n$, $\lfloor\alpha|V|\rfloor$) of the
\State tokens $t$ with highest $L_t$ from $V$ ,
\State where $\alpha \in$ [0, 1] is a hyperparameter
\EndWhile
\State Build final unigram LM $\alpha$ to $S$
\State \textbf{return} $V$, $\theta$
\EndProcedure
\end{algorithmic}
\end{algorithm}

\subsection{Unigram language modeling}
Unigram language modeling (LM) was first applied in NMT in 2018 and compared the performance to that of BPE \cite{kudo2018subword}. Unigram language modeling algorithm takes a set of strings $S$ and aims to create a vocabulary $V$ with a target size of $n$. It starts by initializing $V$ with all substrings occurring more than once in $S$ (without crossing words). The algorithm then iteratively prunes the vocabulary by estimating the token 'loss' $L_t$ for each token $t$ in $V$ using the unigram language model $\theta$. The tokens with the highest $L_t$ values are removed from $V$ until its size reaches the target vocabulary size $n$. Finally, the algorithm fits the final unigram language model $\theta$ to $S$ and returns $V$ and $\theta$ as the resulting vocabulary and language model, respectively.

\subsection{Related work}
The choice of subword units for acoustic modeling can depend on settings such as high-variability spontaneous speech, noisy environment, low-resource scenario, or cross-lingual speech recognition \cite{livescu2012subword}. Different subword modeling techniques have been explored in numerous studies including improved word boundary marker in weighted finite state transducer (WFST)-based decoder for Finnish, Estonian \cite{smit2017improved}, pronunciation-assisted subword modeling (PASM) \cite{8682494}, acoustic data-driven subword modeling (ADSM) \cite{zhou2021acoustic}, among others. While both PASM and ADSM are reported to outperform BPE-based modeling for ASR, these two approaches are evaluated with only a morphologically poor language English. Thus, it is uncertain how different subword modeling approaches will work for morphologically rich languages.

More recently, phone-based BPE has been introduced for multilingual speech recognition \cite{yusuyin2023investigation}. However, in a monolingual setting, PBPE obtains similar performance compared to BPE while both BPE and PBPE outperform phone and character-based modeling.

In recent years, Bengali ASR research has primarily focused on addressing the scarcity of datasets through resource development initiatives \cite{kjartansson2018crowd, ahmed2020preparation, kibria2022bangladeshi, rakib2023oodspeech}. \citet{sadeq-etal-2020-improving} addressed the challenge of manual annotation in training ASR systems by proposing a semi-supervised approach for Bangla ASR, leveraging large unpaired audio and text data encoded in an intermediate domain with a novel loss function. \citet{samin2021deep} evaluated the LB-ASRTD corpus \cite{kjartansson2018crowd}, a large-scale publicly available dataset comprising 229 hours, utilizing deep learning-based methods and performing a character-wise error analysis. While earlier studies on Bengali ASR involved phone-based segmentation \cite{al2019continuous}, the shift towards end-to-end ASR systems has made character-based models more prevalent \cite{samin2021deep}. Notably, to the best of our knowledge, there has been no study comparing different subword modeling techniques in the Bengali speech processing domain.

\section{Method}
\label{sec:method}
We train a convolutional neural network (CNN) based acoustic model for performing the experiments. We extract 21 mel-frequency cepstral coefficients (MFCCs) from each frame of the input signal and feed it to the CNN. The frame length and stride are 30 ms and 15 ms, respectively. We implement the same CNN architecture following the work of \citet{samin2021deep} except for introducing a batch normalization layer in each convolution block. Moreover, we use 20 convolutional blocks instead of 15. The objective of the acoustic model is to predict the subword units based on the CTC loss criterion given the audio signal \cite{graves2006connectionist}. We choose either character, BPE, or unigram tokens in individual experiments as our subwords.

We do not perform beam search decoding with a language model since it can have an impact on the final result. The goal of this study is to investigate different subword-based acoustic modeling for a morphologically rich language Bengali, so we exclude the language modeling part. Therefore, greedy search decoding is used to generate the output tokens.

\begin{table*}[t]
\caption{Six acoustic models are trained with three types of subword units such as character, unigram, and BPE. For BPE segmentation, 500, 1K, 2K, and 3K target tokens are fixed in separate experiments. The models are trained with a CNN architecture on the Bengali SUBAK.KO train set and evaluated on the eval sets of SUBAK.KO, LB-ASRTD, and SHRUTI. TERs (\%) and WERs (\%) are reported.}\label{table_classifier}
\vspace{0.5cm}
\begin{center}
\begin{tabular}{cccccccc}
\toprule
\multirow{2}{*}{\textbf{Token type}} & \multirow{2}{*}{\textbf{\# tokens}} & \multicolumn{2}{c}{\textbf{SUBAK.KO eval}} & \multicolumn{2}{c}{\textbf{LB-ASRTD eval}} & \multicolumn{2}{c}{\textbf{SHRUTI eval}}\\[0pt]
\cmidrule(lr){3-4}
\cmidrule(lr){5-6}
\cmidrule(lr){7-8}
 &  & \textbf{TER} & \textbf{WER} & \textbf{TER} & \textbf{WER} & \textbf{TER} & \textbf{WER}\\[0pt]
\midrule
Character & 73 & 5.41 & 18.89 & 27.14 & 66.44 & 14.31 & 46.34\\
Unigram & 1000 & 6.03 & 16.86 & 30.32 & 66.07 & 15.97 & 44.40\\
BPE & 500 & 5.71 & 17.11 & 28.15 & 64.28 & 14.61 & \textbf{42.80}\\
BPE & 1000 & 5.97 & 16.65 & 29.32 & \textbf{63.80} & 15.97 & 43.75\\
BPE & 2000 & 6.19 & 16.17 & 31.99 & 66.38 & 17.36 & 44.58\\
BPE & 3000 & 6.34 & \textbf{15.63} & 34.11 & 66.46 & 18.84 & 45.77\\[2pt]
\bottomrule
\end{tabular}
\end{center}
\end{table*}

\section{Experiment setup}
\label{sec:experiment_setup}

We implement CNN-based acoustic models using the Flashlight toolkit \cite{kahn2022flashlight}. We train our CNNs using SUBAK.KO, an annotated Bangla speech dataset \cite{kibria2022bangladeshi}. SUBAK.KO is mostly a read speech corpus with 229 hours of read speech and only 12 hours of broadcast speech. We use the same 200-hour long training set, 20-hour long development (dev) set, and 20-hour long evaluation (eval) set following \citet{kibria2022bangladeshi}. Using standard train, dev, and eval sets enables us to compare our strategy to those of the past. For a comprehensive evaluation, we use a 20-hour subset of the large Bangla automatic speech recognition training data (LB-ASRTD) and the 20-hour long full SHRUTI corpus \cite{kjartansson2018crowd, das2011bengali}. Our SUBAK.KO-based ASR model encounters OOV words from LB-ASRTD and SHRUTI out-of-distribution data. Therefore, cross-evaluation assures a more reliable evaluation of various subword modeling algorithms.

We train baseline ASR systems using character and unigram tokens, subsequently contrasting their performance with BPE-based ASR systems. For character-based modeling, we simply use Python programming language to segment a word into individual characters. To build the BPE and unigram-based lexicons, we use the Sentence-piece library \cite{kudo2018sentencepiece}. For unigram language modeling, a fixed token size of 1000 is used. As for BPE, we develop four acoustic models with 500, 1K, 2K, and 3K tokens and compare the results. SUBAK.KO train set is used as a text corpus to generate the BPE and unigram tokens.

For evaluating ASR models, we employ standard metrics such as the word error rate (WER) and the token error rate (TER). Eight graphics processing units (GPUs) with only 12 gigabytes of virtual random access memory each are used to train the models.

\section{Results \& Discussion}
\label{sec:results}

Table \ref{table_classifier} presents the WERs and TERs for various subword modeling types and token sizes. On all three evaluation sets, BPE-based acoustic modeling outperforms both character-based and unigram-based modeling in terms of WERs. Unigram modeling achieves lower WERs than character-based segmentation. With the same number of tokens (1000 tokens), unigram modeling cannot surpass the BPE-based approach when dealing with both in-distribution (SUBAK.KO eval) and out-of-distribution (LB-ASRTD and SHRUTI eval) data.

The number of generated tokens is proportional to the number of BPE merge operations. As we increase the number of BPE tokens, the WERs continue to decrease on the SUBAK.KO eval set. Nevertheless, acoustic models trained with 1000 BPE tokens and 500 BPE tokens achieve reduced WERs on the LB-ASRTD and SHRUTI eval sets, respectively. Notably, BPE tokens are generated utilizing the SUBAK.KO train corpus. This implies that as the number of BPE tokens increases, the model becomes overfit on its eval set while performing poorly on the out-of-distribution data. Thus, a target BPE token size of 500 or 1000 is found to be suitable to achieve better generalizability for ASR. This finding conforms to the work of \cite{gutierrez2023languages}, indicating that morphologically rich languages necessitate fewer BPE merges, leading to a reduced count of BPE tokens. Also, a higher number of BPE merge operations tends to generate longer-length tokens, which resemble words and present similar bottlenecks of word-based tokenization.

With regard to TERs, character-based acoustic modeling exhibits lower error rates than the other two approaches. Furthermore, with the BPE tokens, the TERs tend to continually increase when we increment the number of BPE tokens. Each character represents a token of length one, while BPE tokens undergo multiple merge operations, resulting in longer token lengths. We argue that predicting a token with a shorter length is a comparatively easier task for the acoustic model in contrast to mapping input signal frames to a token with a longer length. This discrepancy can contribute to higher TERs for unigram and BPE-based modeling. However, if a model can accurately predict the long-length tokens of a word, it increases the probability of achieving a lower WER because the number of unigram/BPE tokens in a word is typically fewer than the number of characters in that word.

\begin{figure}
\begin{tikzpicture}
    \begin{axis}[   
        ybar=0.15cm,
        bar width=0.15cm,
        width=8cm,
        height=7cm,
        ymin=0,
        ymax=100,
        yticklabel style={font=\scriptsize},
        xtick=data,
        xticklabels={40 hours, 80 hours, 120 hours, 160 hours, 200 hours},
        xticklabel style={yshift=0ex, rotate=0, font=\scriptsize},
        x label style={at={(axis description cs:0.5,-0.10)},anchor=north, font=\small},
        y label style={font=\small},
        xlabel={Training corpus size},
        ylabel={Word error rate (\%)},
        major x tick style={opacity=0},
        minor x tick num=1,
        minor tick length=1ex,
        ymajorgrids=true,
        legend style={
            at={(0.50,1.04)},
            anchor=south,
            column sep=0.5ex,
            legend columns=-1,
            font=\scriptsize,
        },
    ]
    \addplot [style={blue, fill=blue}] coordinates {
        (1,38.3)
        (2,26.0) 
        (3,21.0)
        (4,18.4) 
        (5,16.6)
    };
    \addplot [style={red, fill=red}] coordinates {
        (1,83.7) 
        (2,73.1) 
        (3,69.8)
        (4,65.8) 
        (5,63.8)
    };
    \addplot [style={green, fill=green}] coordinates {
        (1,63.7) 
        (2,53.6) 
        (3,48.9)
        (4,45.7) 
        (5,43.8)
    };
    \legend{SUBAK.KO eval,LB-ASRTD eval, SHRUTI eval};
    \end{axis}
\end{tikzpicture}
\centering
\caption{Acoustic models are trained with 1000 BPE tokens on five different SUBAK.KO train subsets (e.g. 40 hours, 80 hours, 120 hours, 160 hours, and 200 hours). SUBAK.KO, LB-ASRTD, and SHRUTI eval sets are used to report the WERs (\%)}
\label{fig:training_hours}
\end{figure}
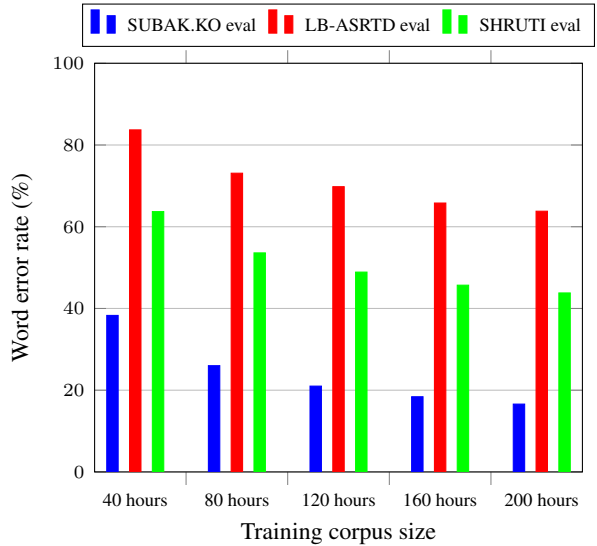

Figure \ref{fig:training_hours} demonstrates the performance of BPE-based subword modeling with various train corpus sizes. We can observe the positive impact of increasing the amount of acoustic model training data on all three evaluation sets for the BPE-based approach. However, it is worth noting that BPE-based speech recognition systems can still achieve satisfactory performance even in low-resource scenarios.

\section{Conclusion \& Future Work}
\label{sec:conclusion}
In this work, we determine the number of BPE merges in the context of ASR for a morphologically rich language - Bengali and provide intriguing insights into the relationship between BPE merge operations and ASR performance in the presence of OOV words. Furthermore, we provide a comparative analysis for three subword modeling approaches including characters, unigrams, and BPE for ASR. Our empirical study suggests that BPE is a better choice for subword modeling than characters and unigram tokens. Additionally, through cross-dataset evaluation, we find that targeting a token size of approximately 500 or 1000 yields improved generalization and robustness, while excessively high numbers of BPE tokens can result in overfitting. This outcome corresponds with prior linguistic research, suggesting that morphologically rich languages demand fewer BPE merges to yield highly productive BPE tokens, as discussed in \cite{gutierrez2023languages}.

There are several potential directions for future research. Firstly, instead of generating BPE tokens from the text files of SUBAK.KO train set, a large-scale text corpus could be constructed specifically for this purpose, enabling the generation of BPE tokens from a more extensive and diverse dataset. We hypothesize that this approach could yield superior BPE representations, resulting in enhanced robustness across out-of-domain data, particularly for challenging morphologically rich languages. Secondly, we aim to explore additional morphologically rich languages from diverse language families, as well as languages like English that lack complex morphology, to further evaluate the effectiveness of subword modeling approaches. Lastly, although our study benchmarks convolutional neural networks (CNNs), it would be valuable to investigate state-of-the-art transfer learning algorithms, such as wav2vec 2.0 \cite{baevski2020wav2vec} and HuBERT \cite{hsu2021hubert}, to determine if BPE subword modeling remains effective with these architectures.

\bibliography{custom}
\bibliographystyle{acl_natbib}

\end{document}